\documentclass{article}

\usepackage{arxiv}

\usepackage[utf8]{inputenc} % allow utf-8 input
\usepackage{hyperref}       % hyperlinks
\usepackage{booktabs}       % professional-quality tables
\usepackage{amsfonts}       % blackboard math symbols
\usepackage{nicefrac}       % compact symbols for 1/2, etc.
\usepackage{microtype}      % microtypography
\usepackage{lipsum}

\usepackage{graphicx}
\usepackage{amssymb}
\usepackage{amsmath}
\usepackage{subfig}
\usepackage{listings}
\usepackage{times}
\usepackage{latexsym}
\usepackage{url}
\usepackage[ruled,linesnumbered]{algorithm2e}
\usepackage[T3,OT2,T1]{fontenc} 
\usepackage[utf8]{inputenc}
\usepackage{caption}
\usepackage[noenc]{tipa}
\usepackage{tikz}
\usepackage{multirow} %for tables in 5.X
\newcommand{\argmin}{\mathop{\mathrm{argmin}}}
\usepackage{amsthm}
\theoremstyle{definition}
\newtheorem{definition}{Definition}

\usepackage{enumitem}

\title{Altruist: Argumentative Explanations through Local Interpretations of Predictive Models}

\author{
  Ioannis Mollas\\
  Aristotle University of \\Thessaloniki, 54636, Greece\\
    \texttt{iamollas@csd.auth.gr}\\
     \And
       Nick Bassiliades\\
  Aristotle University of \\Thessaloniki, 54636, Greece\\
             \texttt{nbassili@csd.auth.gr}\\
             \And
               Grigorios Tsoumakas\\
  Aristotle University of \\Thessaloniki, 54636, Greece\\
             \texttt{greg@csd.auth.gr}\\

}

\begin{document}
\maketitle

\begin{abstract}
Explainable AI is an emerging field providing solutions for acquiring insights into automated systems' rationale. It has been put on the AI map by suggesting ways to tackle key ethical and societal issues. Existing explanation techniques are often not comprehensible to the end user. Lack of evaluation and selection criteria also makes it difficult for the end user to choose the most suitable technique. In this study, we combine logic-based argumentation with Interpretable Machine Learning, introducing a preliminary meta-explanation methodology that identifies the truthful parts of feature importance oriented interpretations. This approach, in addition to being used as a meta-explanation technique, can be used as an evaluation or selection tool for multiple feature importance techniques. Experimentation strongly indicates that an ensemble of multiple interpretation techniques yields considerably more truthful explanations.
\end{abstract}

% keywords can be removed

\section{Introduction}
\label{introduction}

While we witness a revolutionary adoption of Artificial Intelligence (AI) systems in our everyday activities, it is noticeable that many of them advance through Machine Learning (ML). As a result of the development of AI and ML, several ethical problems affecting our society have arisen, and thus the fields of Explainable AI (XAI) and Interpretable ML (IML) have emerged. Specifically, IML promises approaches for identifying discrimination phenomena in ML models~\cite{amazonBias,propublica}, as well as compliance of industry to legal frameworks~\cite{gdpr}. Eventually, ML practitioners and researchers, developing stronger and more accurate models through IML, will be able to understand and explain their tasks and even identify issues, for example biases in a model, that would otherwise remain undetected.

Techniques for IML can be classified as {\em global}, exposing the entire logic of a model, and {\em local}, aiming to explain a single prediction of a model~\cite{adadi}. Moreover, when an interpretation technique can be applied indifferently to any ML model, we speak of a {\em model-agnostic} technique, while when it is only applicable to a specific model, we call it {\em model-specific}. {\em Feature Importance} (FI) interpretation techniques calculate the influence of each feature to the prediction, either at a global or local level. Methods such as LIME~\cite{lime} (model-agnostic) and GradCam~\cite{gradCam} (model-specific for neural networks) are two FI local techniques.

Argumentation concerns the study of how conclusions can be reached through a logical chain of reasoning, that is, claims based, soundly or not, on premises~\cite{argBook}. Based on this concept, different kinds of Argumentation Frameworks ($AF$) have been designed. IML and argumentation share the same goal of persuading someone to accept the validity of a decision. Many approaches that combine explanation and argumentation towards interactive dialogues, do so in a theoretical way. Whether explanations are arguments or not is a matter of debate in the philosophy of science. An interesting view discriminates between arguments and explanations, provided that arguments are used to justify something in doubt, while explanations are used to express an interpretation of something that is incomprehensible~\cite{bex2016combining}.

Several techniques used to acquire interpretations from ML models are approximations of the real interpretation, which is unknown. Therefore, their validity is questionable. This paper introduces {\em Altruist}; a preliminary method for transforming FI interpretations of ML models into insightful and valid explanations using argumentation based on classical logic. Altruist extracts the local maximum truthful part of an interpretation, providing reasons for the truthfulness justification. Given multiple interpretations, Altruist can work as a meta-explanation technique, as well as a tool to easily choose between $X$ number of different interpretation techniques. Argumentation works as an explanation to this whole process. Altruist's power in recognising untruthful parts of interpretations and usefulness as a meta-explanation technique is demonstrated through experimentation on ML models trained on tabular data. Altruist has innate virtues such as truthfulness, transparency and user-friendliness that characterise it as an apt tool for the XAI community.

\section{Background}
\label{s:argr}

In this section, basic notions of argumentation and IML are introduced. A lot of frameworks have been developed in the area of argumentation, with a similar well-defined mathematical foundation~\cite{argBook,vassiliades2021argumentation,dmollinSBS20}. Abstract~\cite{dung1995argumentation}, Bipolar~\cite{cayrol2005acceptability} and Classic Logic-based~\cite{Besnard2009} argumentation are some of the most well-known $AF$s.

Argumentation based on Classical Logic ($CL$) concerns a framework defined exclusively with logic rules and terms. A sequence of inference to a claim is an argument in this framework. Specifically, an argument is a pair $\langle \Phi, \alpha \rangle$ such that $\Phi$ is consistent ($\Phi \nvdash \perp$), $\Phi \vdash \alpha$, and $\Phi$ is a minimal subset of $\Delta$ (a knowledge base), which means that there is no $\Phi'\subset\Phi$ such that $\Phi' \vdash \alpha$. $\vdash$ represents the classical consequence relation. In this framework counterarguments, the defeaters, are also defined. $\langle \Psi, \beta \rangle$ is a counterargument for $\langle \Phi, \alpha \rangle$ when the claim $\beta$ contradicts the support $\Phi$. Furthermore, two more specific notions of a counterargument are defined as \textit{undercut} and \textit{rebuttal}. Some arguments specifically contradict other arguments' support, which leads to the undercut notion. An undercut for an argument $\langle \Phi, \alpha \rangle$ is an argument $\langle \Psi, \neg(\phi_1 \land \dots \land \phi_n)\rangle$ where $\{\phi_1, \dots, \phi_n\} \subseteq \Phi$. If there are two arguments in objection, we have the most direct form of dispute. This case is represented by the concept of a rebuttal. An argument $\langle \Psi, \beta \rangle$ is a rebuttal for an argument $\langle \Phi, \alpha \rangle$ if $\beta\leftrightarrow\neg\alpha$.

Argumentation begins when an initial argument is put forward, and some claim is made. This leads to an argumentation tree \textit{Tr} with root node the initial argument. Objections can be posed in the form of a counterargument. In \textit{Tr}, these are represented as children of the initial argument. The latter is addressed in turn, ultimately giving rise to a counterargument. Finally, a judge function decides if a \textit{Tr} is rather Warranted or Unwarranted, based on marks assigned to each node as either undefeated \textit{U} or defeated \textit{D}. A \textit{Tr} is judged as Warranted, \textit{Judge(\textit{Tr}) = Warranted}, if \textit{Mark($A_r$) = U}, where $A_r$ is the root node of \textit{Tr}, is undefeated. For all nodes $A_i \in$ \textit{Tr}, if there is a child $A_j$ of $A_i$ such that \textit{Mark($A_j$) = U}, then \textit{Mark($A_i$) = D}, otherwise \textit{Mark($A_i$) = U}.

The ability of ML models to give users insights into their structure and decisions is known as interpretability feature~\cite{adadi}. Decision trees, rule-based models and linear models are inherently (intrinsically) interpretable, and they can provide both local and global information. Model-agnostic interpretation techniques such as Anchors~\cite{RibeiroAnchors:Explanations} choose to lay down rules for local explanation, as well as several model-specific approaches, providing global (e.g.~\cite{domingos}) or local (e.g.~\cite{moore}) explanations. Model-specific techniques, such as GradCam, locally interpret neural networks, for image recognition or object detection tasks, and present their findings using saliency maps (heatmaps) or bounding boxes. LIME, a model-agnostic local-based interpretation technique, introduces a variation of its main algorithm focusing on such image-oriented models, providing saliency maps as explanations as well.

The aforementioned techniques are also able to provide their explanations in the form of FI, when the input data are tabular or textual. In this family of model-agnostic interpretation techniques for any ML model there are the global-based variants of feature Permutation Importance (PI) methods~\cite{RForests}, as well as SHAP~\cite{shap}, another method for calculating the importance of a feature for both global and local aspects.

\section{Related Work}
\label{sec:relwork}

In this work, we are attempting to a) evaluate IML techniques, b) select or ensemble the best among them, and c) explain the entire process of evaluating and selecting IML techniques using argumentation. Therefore, in this section we present the related work.

There are a few works trying to connect IML and argumentation. AA-CBR is a hybrid method that combines case-based reasoning with abstract argumentation. In AA-CBR there are cases, where each one is a set of features and an outcome, and the objective is to predict the outcome of new cases~\cite{AACBR}. ANNA attempts to solve classification problems by using neural auto-encoders for feature selection, AAF for arguments' generation, and AA-CBR for prediction tasks, offering argument sub-graphs as explanations~\cite{ANNA}. ABML~\cite{movzina2007argument} is a technique inspired by the CN2 algorithm, incorporating arguments into the learning process, aiming to reduce the space of the hypotheses. This, along with the interpretable essence of CN2, explanations can be given in the form of arguments. Another technique, CleAr~\cite{carstens2017using}, is a classification technique that incorporates knowledge in the form of arguments, using a variation of the bipolar argumentation framework, with supervised learning applied in computational linguistics. There are, as of yet, no works that use argumentation to explain the process of evaluating and selecting FI approaches.

Regarding IML, a critical domain within this research area is related to the evaluation metrics, which are available for benchmarking and selection processes. There are a few metrics, such as \emph{fidelity} or the \emph{number of non-zero weights}, also known as \emph{complexity}~\cite{adadi}, that are the most common options for assessing FI-based interpretation techniques. Nonetheless, these metrics cannot reflect the effectiveness of the evaluated approaches, as they cannot capture the quality of explanation. At the same time, when there are available ground truth explanations, also known as \emph{rationales}~\cite{rationalees}, we can measure the performance of FI interpretation techniques using the Area Under the Precision-Recall Curve (AUPRC), Intersection-Over-Union (IOU), and F$_1$ score. However, we can only assume that humans can provide meaningful \emph{rationales} if we consider the possible inner bias of each annotator~\cite{tan2021diversity}. Based on the concept of \emph{rationales}, the ERASER benchmark, exclusively for NLP tasks, defines two metrics: \textit{comprehensiveness} and \textit{sufficiency}~\cite{eraser}. The former evaluates the interpretation observing the prediction by removing the rationales from the input, while the latter by retaining only the rationales in the input.

\begin{figure*}[th]
    \centering
    \includegraphics[width=0.95\textwidth]{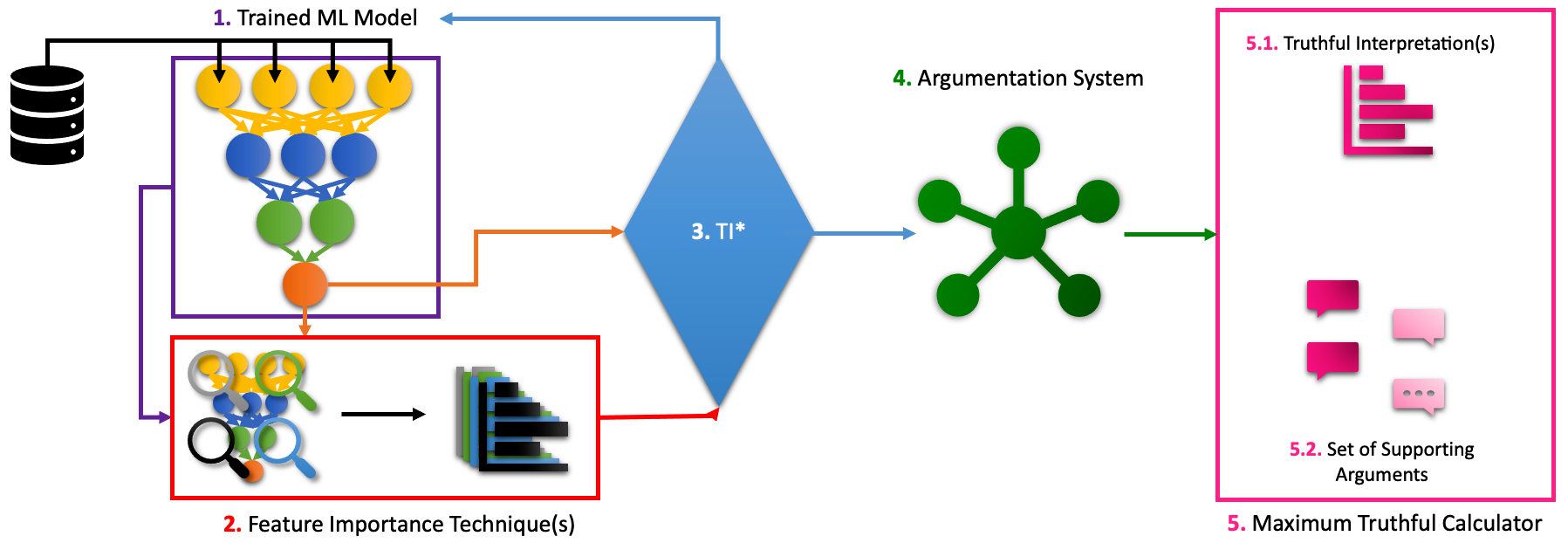}
    \caption{Altruist's flow chart. *TI: Truthfulness Investigator}
    \label{fig:Altruist}
\end{figure*}

\emph{Faithfulness}~\cite{faithfulness}, a similar metric to \emph{comprehensiveness} and \emph{sufficiency}, does not require annotated rationales to measure the quality of the provided interpretations. \emph{Faithfulness} is applicable to FI-based techniques tested on models handling textual data, evaluating if the positive importances of a document's sentences are true. Dealing with the drawbacks of \emph{faithfulness}, \emph{infidelity} defines different ways to create perturbations to test the \emph{faithfulness} of an interpretation~\cite{infidelity}. However, \emph{infidelity} is examined in the context of image data, and its output ranges in $[0,+\infty)$, making comparisons between various algorithms and datasets more challenging.

There are few works focusing on aggregating and ensembling multiple interpretation techniques. A recent work focused on \emph{aggregating} FI interpretations to eventually produce interpretations with low sensitivity, high faithfulness, and low complexity~\cite{aggr1}. Specifically, for a given instance, they identify a set of neighbours. They extract interpretations for all of them, and then they aggregate them. Another study, focusing on image classification tasks and based mainly on the stability metric (also known as robustness), aggregates both FI techniques and rule-based approaches by averaging explanation vectors~\cite{bobek2021towards}.

Finally, recent work presented a method for providing a human-centric method of justifying a task-related prediction~\cite{biranM17}. They examine the feature importance (weights) of features using an ML model and an interpretability technique. They provide each feature with a narrative role descriptor depending on its importance and impact on the prediction. Furthermore, they create core messages based on those narratives, using Natural Language Generation (NLG) to generate simple, brief, qualitative, and intuitive justifications for the predictions. One of the work's limitations is that it is applied to linear ML models, which are always correct. The problem stems from the assumption that the linear model's coefficients, or, in the case of a black-box model, the weights generated via techniques such as LIME, are correct. In the second scenario, the weights could be incorrect. In this work, we use Altruist to determine the truthfulness of the weights. Because Altruist acts on a lower level, this work can be used after Altruist.

\section{Altruist}
Altruist aims to tackle mainly the untruthfulness of FI-based approaches, using logic-based argumentation. Altruist's ultimate objective is to supply the largest truthful subset of an instance's interpretation, acting as a curious user who tests various inputs to evaluate the given interpretation. Additionally, Altruist can provide reasons (i.e., arguments) to justify why this maximum subset is truthful, as well as why the features excluded from the set are untruthful. Truthfulness is introduced in Section~\ref{ATIS}. Finally, it can be used as a unified selection or evaluation tool between multiple FI techniques using local perturbations of instances, influenced by faithfulness and infidelity, while it can also be used as a meta-explanation ensemble of FI techniques, similar to aggregation of FIs.

In the present work, Altruist is applied to ML models trained with tabular data. Altruist can indeed be applied to textual, time-series, and even image data in future research. The methodology consists of 5 components (Figure~\ref{fig:Altruist}). The first component includes the ML model, the second component is the interpretation technique(s), the third component is the truthfulness investigator (TI), the fourth component is the argumentation system, while the fifth component offers the final interpretation.

\subsection{Machine Learning Model}
The first component of the technique is the ML model to be interpreted. The ML model could be any model trained on tabular data, that is able to provide continuous values as output, e.g. a classification model which can output probabilities or regression model. This component is referred to as \textit{M}, which is trained on the input dataset $D=[x_i,\dots,x_N]$, which contains $N$ instances with $|F|$ features, where $F=[f_1, \dots, f_{|F|}]$. Each $x_i \in D$ instance has a set of values for the $|F|$ features $x_i=[v_{1,i}, \dots,v_{{|F|},i}]$. The output of this component is the prediction for an instance $x_i$, for example $P_{M}(x_i)=y_i$ in a supervised learning model.

\subsection{Feature Importance Technique(s)}
This next component concerns the interpretation technique(s) and is highly correlated with the \textit{M} component. The interpretation technique(s) must fall within the category of feature importance, and must therefore provide explanations in the form of sets of features accompanied by an indicator of importance. 

At this point, we assume that various FI approaches produce weights with a monotonic notion. This intuitive assumption is used by the majority of global and local FI techniques. This means that an end-user can expect monotonic behaviour from the prediction model when altering a feature based on the weight provided by the global or local FI technique. However, there are FI techniques, like as SHAP, that do not presuppose monotonicity but are perceived as such by the end user.

Such techniques may be global or local, as well as model-agnostic or model-specific. The output of this component, given a specific $x_i$, and the \textit{M} component, is denoted as \textit{Z}$=[z_1,...,z_{|F|}]$, where $z_j \in \mathbb{R}$. It is possible to have multiple interpretation techniques to let Altruist choose the best (most truthful) interpretation. Then, for $T$ different techniques, one would have \textit{Z}$_t$, where $t \in [1,T]$.

\subsection{Truthfulness Investigator}
\label{ATIS}
The third component of this methodology is the Truthfulness Investigator (TI). For a specific $x_i$, this component takes as input the FI interpretation(s) of $x_i$ from the previous component. Based on the faithfulness and infidelity metrics introduced in Section~\ref{sec:relwork}, TI investigates locally (e.g., around an instance) if the FIs are truthful or not. Altruist mimics human behaviour with this component. When an end-user receives an explanation, either local or global, they can alter values to see if the prediction changes. TI exhibits the same behaviour, but only locally, i.e., the alterations it performs are relatively small.

\begin{algorithm}[ht]
 \KwIn{$value, distribution\_of\_feature$}
 \KwOut{$value^-, value^+$}
    $min, max, mean, std \gets extract(distribution\_of\_feature)$ \\
    $noise \gets| mean - gaussian\_noise(mean, std) |$\\
    $value^- \gets value - noise$\\
    $value^+ \gets value + noise$\\
   \lIf{$value^- < min$}{
    $value^- \gets min$
  }
  \lIf{$value^+ > max$}{
      $value^+ \gets max$
  }
   \KwRet{$value^-, value^+$} 
 \caption{Process of determining the alternative values for a feature}
 \label{alg:altvalues}
\end{algorithm}

Initially, TI measures the distribution -- the domain -- of each feature in the training set. Based on the distribution of each feature, 2 alterations are selected, resulting in $2\times|F|$ alterations of $x_i$, which then evaluate the truthfulness of each feature $f$. These two alterations $v_{f,i}'$, for each feature, are performed by adding and removing a small Gaussian noise ($noise$) on the instance's value $v_{f,i}' = v_{f,i} \pm noise$. Therefore, TI creates $2\times|F|$ neighbours very close to the instance whose interpretation we are investigating, to evaluate locally the truthfulness of the interpretation. More details are presented in Algorithm~\ref{alg:altvalues}. For example, two alterations to the value of the `\textit{age}' feature ($v_{age,i} = 24$) for a particular instance $i$ will be ($v_{age,i}' = 24\pm noise$). In this way, we simulate a curious user who, given a FI explanation, makes small modifications to the values originally provided to the system locally to verify the validity of the explanation. To formulate this process, we introduce the definitions of importance, alteration, and expectation, as well as truthfulness.

\begin{definition}
\label{th:imp}
The importance assigned to a feature can be IMP $\in$ \{$1=$ Positive ($z_i>0$), $-1=$ Negative ($z_i<0$), $0=$ Neutral ($z_i=0$)\}.
\end{definition}

\begin{definition}
\label{th:alt}
The alteration of the value of a feature can be ALT $\in$ \{$1=$ Increasing ($v_{j,i}'>v_{j,i}$), $-1=$ Decreasing ($v_{j,i}'<v_{j,i}$)\}, where $v_{j,i}'$ the altered value.
\end{definition}

\begin{definition}
\label{th:exp}
The expected behaviour of an \textit{M} component can be EXP $\in$ \{$1=$ Increasing ($P_{M}(x_i')>P_{M}(x_i)$), $-1=$ Decreasing ($P_{M}(x_i')<P_{M}(x_i)$), $0=$ Remaining Stable ($|P_{M}(x_i') - P_{M}(x_i)|\leq\delta$)\}, where $x_i'$ the instance with the altered value, while tolerance $\delta$ is defined either manually by the user or is set to a default value ($0.01$).
\end{definition}

\begin{table}[ht]
\centering
\resizebox{0.37\textwidth}{!}{%
\begin{tabular}{cc|c|c|c|c|c|c|c}
\cline{3-9}
                                           &    & \multicolumn{3}{c|}{1} & \multicolumn{3}{c|}{-1} & \multicolumn{1}{c|}{\textbf{ALT}} \\ \cline{3-9} 
                                           &    & 1     & 0     & -1     & 1      & 0     & -1     & \multicolumn{1}{c|}{\textbf{EXP}} \\ \hline
\multicolumn{1}{|c|}{} & 1  & t     & u     & u      & u      & u     & t      &                                   \\ \cline{2-8}
\multicolumn{1}{|c|}{{\textbf{IMP}}}                    & 0  & u     & t     & u      & u      & t     & u      &                                   \\ \cline{2-8}
\multicolumn{1}{|c|}{}                     & -1 & u     & u     & t      & t      & u     & u      &                                   \\ \cline{1-8}
\end{tabular}
}
\caption{Truthfulness matrix [(t)ruthful and (u)ntruthful states]}
\label{tab:truth}
\end{table}

\begin{definition}[Truthfulness]
\label{th:tru}
The importance assigned to a feature can be defined as \textit{truthful} when the expected changes to the output of the \textit{M} model $P_{M}(x_i')$ are correctly observed with respect to the alterations that occur in the value of this feature. Thus, for both values of ALT and a given IMP, the IMP$\times$ALT=EXP must be in accordance with the truthfulness matrix (Table~\ref{tab:truth}).
\end{definition}

Therefore, for a positive IMP, these alterations should increase the prediction, for the increased value modification, and decrease the prediction, for the decreased value. For features with negative IMP, the inverse behaviour is expected. If the IMP was neutral, $z_i=0$, we would expect the prediction to remain stable for both the increased and decreased values, $v_{j,i}^{inc}$ and $v_{j,i}^{dec}$, respectively, or to be altered within a very tight range $\delta$, e.g., $0.749$ to $0.750$. Tolerance $\delta$ is defined either manually by the user or is set to a default value. $0.01$ was selected after experimentation as default value because it represents an insignificant alteration of a prediction.

It is worth demonstrating this with an example. For a random instance $x_i$ assigned to class $Y$ with probability $P_{M}(x_i)=0.7$, the feature $f_1$, with a value of $v_{1,i} = 1$, has acquired an IMP $z_1 = 0.5$ (Positive). Altruist seeks to increase and decrease the value of the feature by using Gaussian noise based on its distribution, $v_{1,i}^{inc} = 1.21$ and $v_{1,i}^{dec} = 0.85$. By querying the \textit{M} ML model, it observes the alteration of the model's output. In this example, for the $v_{1,i}^{inc}$ the output was raised to $0.85$, and for the $v_{1,i}^{dec}$ remained stable.

This component does not judge the truthfulness of a feature, but it generates predicates in the output. The following is a specific predicate example that could be generated by this component: ``The model's behaviour by \textit{Increasing} $f_2$'s value is not according to its \textit{Positive} importance''. Such predicates are generated and used as input to the argumentation system and are fully described in the following section. 

The reason why we need the following component is that it is simpler to turn this problem into an argumentation problem and solve it by using logic instead of manually coding it. In fact, using this approach, the results for the selection of features are also justifiable, so we have an all-inclusive transparent method. Finally, using logic and argumentation, the output of this component is a set of natural language arguments that can be easily used by the user, or can be even utilised in a user-chatbot dialogue.

\subsection{Argumentation System}
\label{sec:argsys}

The fourth component, having as input the predicates generated by the TI, is responsible for testing the truthfulness of each feature, determining the set of features that are truthful, and providing explanations on this basis. An AF based on logic is employed to accomplish this. We chose argumentation based on logic because it is more intuitive for the end user than an abstract argumentation framework.

Except for the type of arguments that exist in the framework, we should define the rebuttal and undercut attacks, as well as the \textit{Tr} and a judge function that checks the trust of the whole explanation. The predicates produced by the TI component can formulate atoms of the following types:

\begin{itemize}[leftmargin=0.9em]
  \setlength\itemsep{0.2em}
    \item []$a$: The explanation is untrusted
    \item []$b$: The explanation is trusted
    \item []$c_j$: The importance $z_j$ is untruthful
    \item []$d_j$: The importance $z_j$ is truthful
    \item []$e_{j,ALT}$: The model's behaviour by altering $f_j$'s value is not according to its importance
    %The model's behaviour by ALT $f_j$'s value is not according to its IMP
    \item []$f_{j,ALT}$: The evaluation of the alteration of $f_j$'s value was performed and the model's behaviour was as expected, according to its importance.
    %$f_j$'s value got ALT and evaluated and the model's prediction was EXP as expected
\end{itemize}

%The difference between the last three atoms, are that $d_j$ claims that an importance is truthful having, for example, positive importance, when the   
%Για αρχή εδώ τα atoms τα χρησιμοποιώ σαν καλούπια. πχ το d μπορεί να είναι άτομα με όλους τους δυνατούς συνδυασμούς. Ωστόσο το προηγούμενο component (TI) θα παράξει predicates που θα έχουν συγκεκριμένα ALT, EXP, etc που θα ανήκουν στο καλούπι d. Τώρα το d με το e έχουν διαφορά στο οτί το d λέει οτί γενικά το importance ειναι truthful και από τις δυο πλευρές, είτε όταν κάνουμε increase είτε όταν κάνουμε decrease το value (to alteration dld), ενώ το e, επικεντρώνεται στο κάθε ένα από αυτά ξεχωριστά, και για αυτό επειδή το ένα από τα 2 μπορεί να είναι σωστό, ενώ το άλλο όχι. Έτσι έστω ένα από τα 2 είναι λάθος και το argument που θα περιέρχει το d θα δέχεται επίθεση.

Based on the aforementioned atoms, we can present the arguments of our AF in the form of $\langle\Phi,\alpha\rangle$, where $\alpha$ is the claim of the argument and $\Phi$ is the support:
\begin{itemize}[leftmargin=0.9em]
  \setlength\itemsep{0.2em}
    \item []$\alpha_1$: $\langle\{a\},a\rangle$
    \item []$\alpha_2$: $\langle\{b, b \to \neg a \},\neg a \rangle$
    %\item []$\alpha_2$: $\langle\{b, b \to \neg a \},\neg a \rangle $
    \item []$\alpha_3$: $\langle\{(c_1 \lor\dots\lor c_j),(c_1 \lor\dots\lor c_j)\to \neg b \},\neg b \rangle $
    \item []$\alpha_{4,j}$: $\langle\{d_j, d_j \to \neg c_j \},\neg c_j \rangle$
    \item []$\alpha_{5,j}$: $\langle\{(e_{j,inc} \lor e_{j,dec}),(e_{j,inc} \lor e_{j,dec}) \to \neg d_j \},\neg d_j \rangle$
    \item []$\alpha_{6,j,ALT}$: $\langle\{f_{j,ALT}, f_{j,ALT} \to \neg e_{j,ALT} \},\neg e_{j,ALT} \rangle$
\end{itemize}

Most arguments are self-explanatory. We only briefly explain $\alpha_3$, which states that if the importance of one feature is untruthful, then the explanation is not trusted, and $\alpha_{5,j}$, which states that if one of the applicable alterations of $f_j$'s value is not according to its importance, then the importance $z_j$ is not truthful. These arguments are explained in Section~\ref{s:exam}. It is important to notice that the first argument, $\alpha_1$, is trivial, provided only to ease conversion of arguments into discussions.

\begin{table}[ht]
\centering
\resizebox{0.4\textwidth}{!}{%
\begin{tabular}{c|c|c|c|c|c|c|}
\cline{2-7}
 & \multicolumn{6}{c|}{\textbf{Types of Arguments}} \\ \cline{2-7} 
\textbf{} & $a_1$ & $a_2$ & $a_3$ & $a_{4,j}$ & $a_{5,j}$ & $a_{6,j,ALT}$ \\ \hline
\multicolumn{1}{|c|}{\textbf{Rebuttals}} & $a_2$ & $a_1$ & - & - & - & - \\ \hline
\multicolumn{1}{|c|}{\textbf{Undercuts}} & - & - & $a_2$ & $a_3$ & $a_{4,j}$ & $a_{5,j}$ \\ \hline
\end{tabular}
}
\caption{Attack relations between arguments}
\label{tab:attacks}
\end{table}

We also define the attack relations between such arguments. We use the special cases of attacks, undercut and rebuttal as discussed in Section~\ref{s:argr}, presenting them in Table~\ref{tab:attacks}.% $atts \in$ [rebuttal, undercut].

\begin{figure}[ht]
    \centering
    \includegraphics[width=0.47\textwidth]{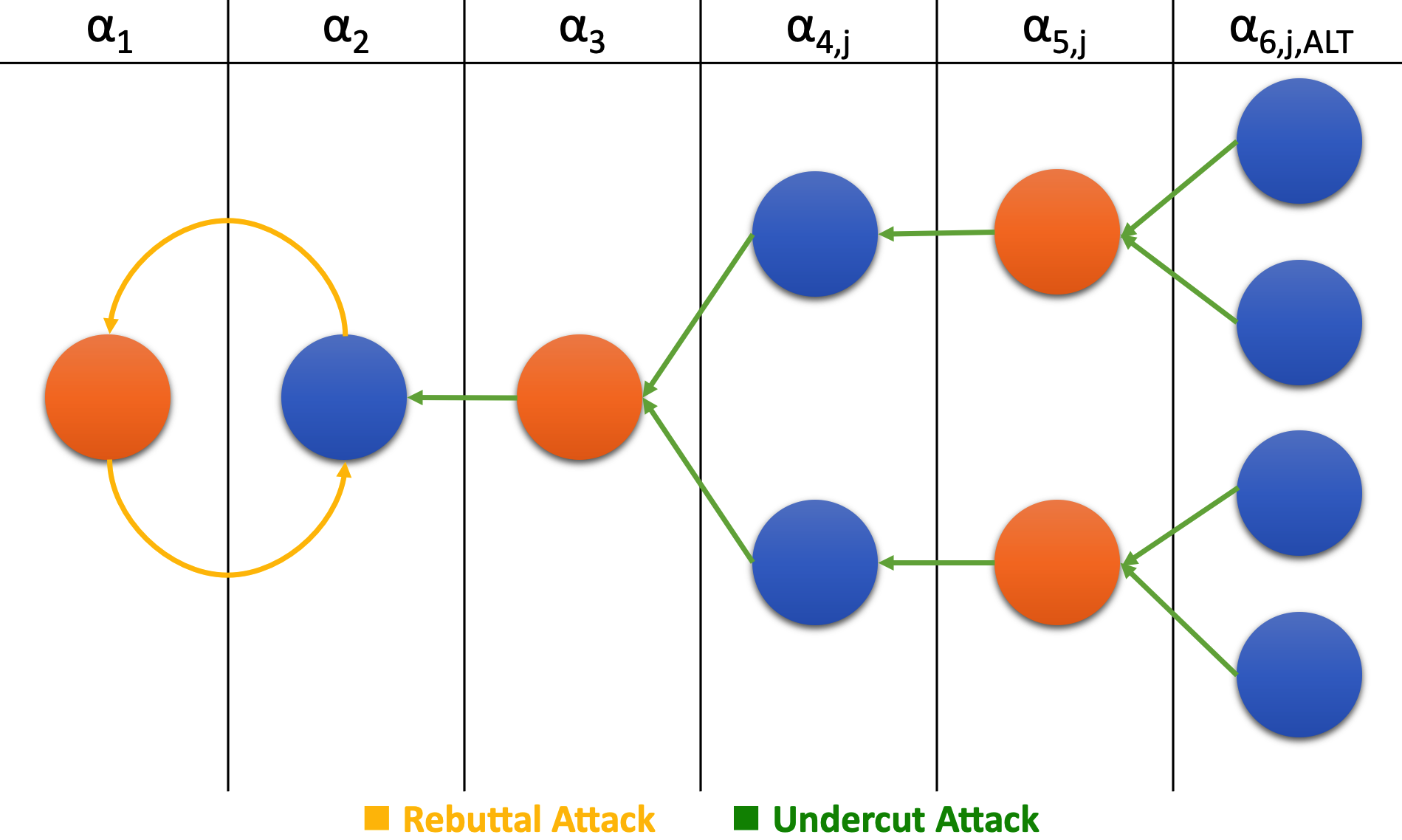}
    \caption{Types of arguments and \textit{Tr}}
    \label{fig:argTree}
\end{figure}

We can proceed to the definition of the argumentation tree \textit{Tr}. A \textit{Tr} begins when an initial argument is presented as a claim and is called root argument. In the form of a counterargument, an objection (or objections) is raised. This is articulated in turn, eventually leading to a counterargument if it is feasible. In Altruist, the root argument is always $\alpha_1$. Thus, the \textit{Tr} is similar to the structure of the tree in Figure~\ref{fig:argTree}. The goal is to decide whether the root argument of this tree is defeated. We use the following judge function:

\[
  J(Tr) =
  \begin{cases}
        \text{Unwarranted} & \text{if } Mark(\alpha_1) = \text{(D)efeated}\\
        \text{Warranted} & \text{if } Mark(\alpha_1)= \text{(U)ndefeated}
  \end{cases}
\]

The root argument ($\alpha_1$) is U if the attacking argument ($\alpha_2$) is \textit{D}. Thus, every argument is marked:

\[
    mark(\alpha_i) = 
    \begin{cases}
        \text{U} & \text{if } Mark(\alpha_j) = \text{D}, \forall \alpha_j \in opp(\alpha_i)\\
        \text{D} & otherwise
    \end{cases}
\]
where the \textit{acceptable} arguments (arguments without opponents/ conflicts) are marked as U, while $opp(\alpha_i)$ are the attacking arguments of $\alpha_i$.

To judge the $Tr$, we utilise a Prolog script, which outputs the arguments in a natural language form. In case the Prolog program judges the root argument $\alpha_1$ as $U$, and therefore the \textit{Tr} as Warranted, this means that one or more features are untruthful. Then, these features are discarded, and by re-examining the \textit{Tr}, we expect to be Unwarranted. The output to the following component is the new reduced interpretation, or interpretations in case of many techniques, which will be \textit{Z}$'=[z^t_1,z^t_2,z^u_3...,z^u_{|F|}]$, where $z^t_i$ the truthful feature importances, and $z^u_i$ the untruthful. An explanatory example is presented in Section~\ref{s:exam}. 

%\begin{figure*}[ht]
%    \centering
%    \includegraphics[width=0.85\linewidth]{newexample.png}
%    \caption{Argumentation tree for a truthful (without red boxes) and an untruthful scenario (with red boxes)}
%    \label{untruthTrees}
%\end{figure*}

\begin{figure*}[ht]
    \centering
    \includegraphics[width=0.87\linewidth]{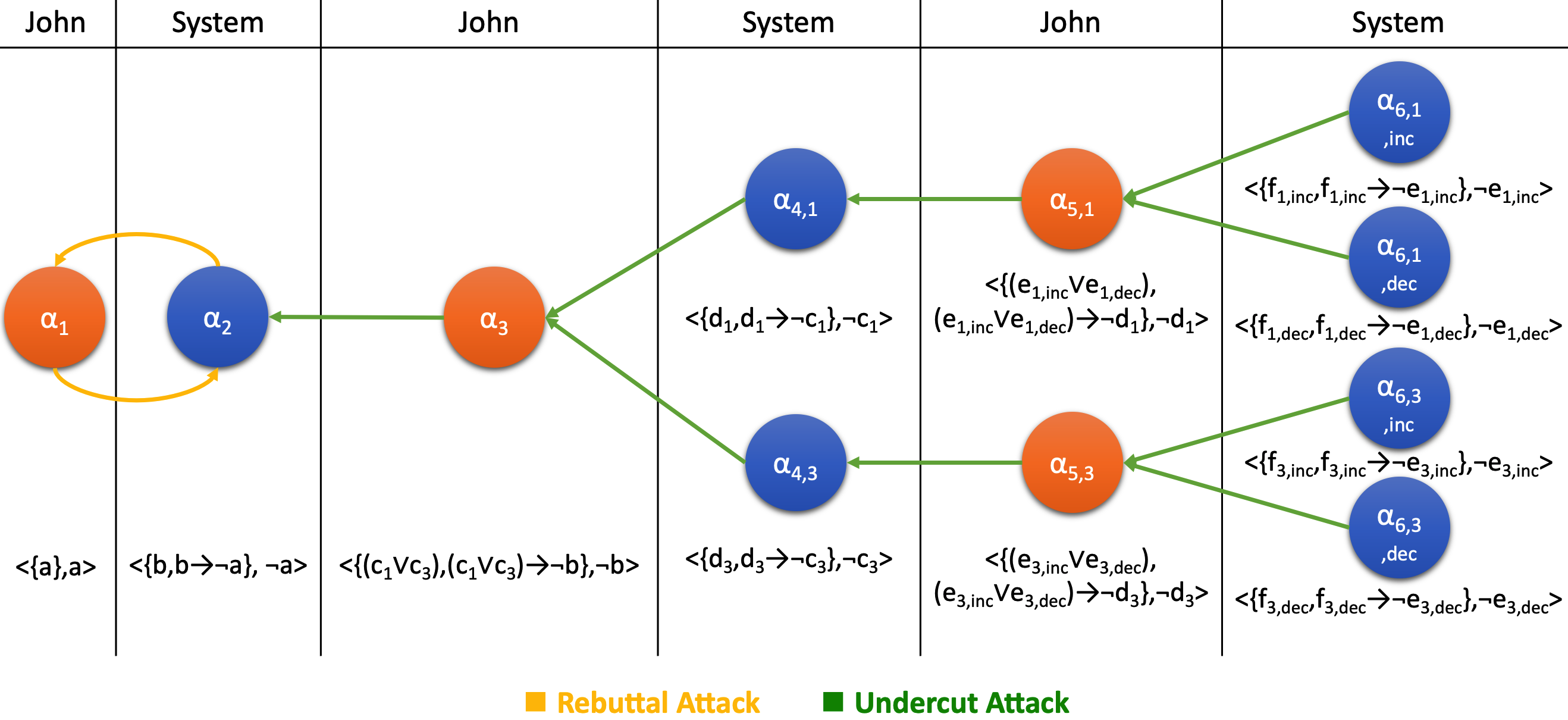}
    \caption{Argumentation tree for the simple example}
    \label{simpleExTrees}
\end{figure*}

\subsection{Maximum Truthful Calculator}

The previous component provides information about which features' importances are truthful and untruthful in the interpretation, for each FI technique if more than one is used. Then, it reforms the interpretation excluding all the untruthful features $z^u_i \in$\textit{Z}$'$. If there are multiple FI techniques, it chooses as the final interpretation the one with the minimum number of untruthful features, $\argmin_t |[z^u_i | z^u_i \in \textit{Z}_t', i \in [0,|\textit{Z}_t'|]]|$, hence the maximum number of truthful features. This interpretation provides richer details, as more features appear, and more accurate results, that can be tested by the end user. Moreover, due to the transparent nature of argumentation, Altruist can explain why a feature is excluded or included. A detailed qualitative experiment takes place in the following Section~\ref{s:exp}. The results of this component could be used by a system designer in a textual format, by converting the arguments into phrases in natural language. The output of Altruist is not intended to be presented directly to end users. A higher-level application, such as a chatbot, is expected to use Altruist.

\subsection{Illustrative Example}
\label{s:exam}
%The paper was fairly well written throughout and the presentation was reasonable. One major issue, in my opinion, however, was the running example. Although it was very welcome and I appreciated the dedication of such a large amount of space towards it, I did not feel like it was very clear at all, despite understanding the approach beforehand. I think it could have been clearer if the example were simpler with more description about the motivations. Further, having the alternative scenario with the omitted feature throughout the example could be very confusing. I think two separate (possibly simpler) examples would have been clearer.

%It would be good to see "age", "weight" and "height" in Figure 3, instead of generic variables c_1, c_2, d_1, d_3 etc. 

To demonstrate the functionality of our methodology, we present a simple but complete example. The example depicts a user interacting with a system via dialogue (e.g., a chatbot) to understand why a prediction is made and to investigate the truthfulness of the provided explanation. We will demonstrate how employing Altruist on top of another interpretation technique, in this case, LIME, can help to prevent presenting misleading information to the end-user by detecting truthful and untruthful feature importance scores. Figure~\ref{simpleExTrees} illustrates this example.

Suppose there is a system solving a classification problem with only three features; Age (\textit{`A'}), Weight (\textit{`W'}) and Height (\textit{`H'}), which predicts the probability of `Author's Paper Approval'. A probability of $[0,0.5)$ means that the paper will be rejected, while a probability of $[0.5,1]$ means that the paper will be accepted. John is a PhD student who is 25 years old, his weight is 62 kg, and his height is 170 cm and asks the system for a prediction. The \textit{M} component of the system predicted a probability of $0.75$ of his paper to be accepted. John is also given an explanation through LIME, corrected by Altruist, suggesting that his \textit{A} is positively influencing ($z_1=0.5$) the probability of his paper to be accepted, while his \textit{H} has neutral influence ($z_3=0$). The actual LIME explanation also included an importance value for \textit{W} ($z_2=0.1$), but Altruist considered it untruthful and chose not to present it.

John is presented with arguments generated by Altruist. The first argument $\alpha_1$ claimed: $a=$``The explanation is not truthful''. John can generally raise this argument to derogate the truthfulness of the explanation. Subsequently, the second argument claims that $b=$``The explanation is truthful''. This argument, $\alpha_2$, is provided by the system, and it is a rebuttal to $\alpha_1$. At this point, we can observe that, as indicated in Section~\ref{sec:argsys}, the argument $\alpha_1$ is trivial, but appears crucial for the discussion's flow.

For each one of the $|F|$ features appearing in the explanation a claim $c_j$, $j \in [0,|F|]$ can be raised by John, stating that ``The $z_j$ is untruthful''. Specifically, two claims are raised $c_1, c_3$, composing argument $\alpha_3$ which is an undercut attack to the argument $\alpha_2$. These claims are $c_1=$``The importance of \textit{A} is untruthful'' and $c_3=$``The importance of \textit{H} is untruthful''. Note here that $c_2=$``The importance of \textit{W} is untruthful'' is not stated by John, as Altruist opted for this feature importance, and never reached John. In the end of this example, the omission of this feature is explained as well. Altruist creates another 2 claims to answer, $d_1=$``The importance of \textit{A} is truthful'', and $d_3=$``The importance of \textit{H} is truthful''. Two arguments $\alpha_{4,1}$ and $\alpha_{4,3}$ are composed undercutting $\alpha_3$.

To prove these claims, John raises four new claims $e_{1,Inc},$ $e_{1,Dec},$ $e_{3,Inc},$ $e_{3,Dec}$. For example, we present two of them: $e_{1,Inc}=$``The model's behaviour by Increasing \textit{A}'s value is not according to its Positive importance'' and $e_{1,Dec}=$``The model's behaviour by Decreasing \textit{A}'s value is not according to its Positive importance''. Each pair of these claims (e.g., $e_{1,Inc}, e_{1,Dec}$) form an argument $\alpha_{5,1}$, which is an undercut attack to $\alpha_{4,1}$. Finally, the last four claims are generated $f_{1,Inc},f_{1,Dec},f_{3,Inc},f_{3,Dec}$; each pair is forming an argument $\alpha_{6,j,ALT}$ which undercuts the argument $\alpha_{5,j}$. These claims are for example: $f_{1,Inc}=$``The evaluation of the alteration (Increased) of \textit{A}'s value is performed and the model's behaviour is as expected (Increased), according to its importance'' and $f_{1,Dec}=$``The evaluation of the alteration (Decreased) of \textit{A}'s value is performed and the model's behaviour is as expected (Decreased), according to its importance''.

\begin{figure*}[ht]
    \centering
    \includegraphics[width=0.87\linewidth]{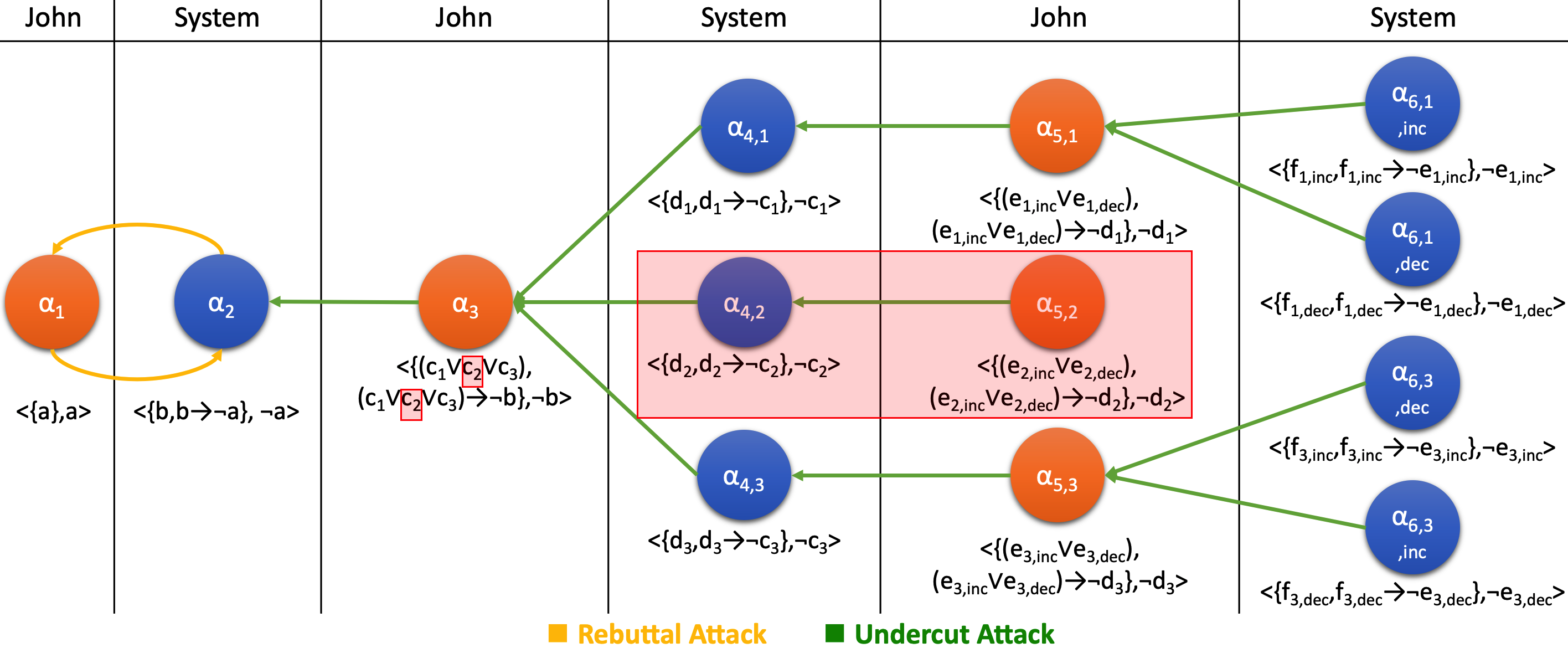}
    \caption{Argumentation tree for an untruthful scenario}
    \label{untruthTrees}
\end{figure*}

However, if any of these last arguments are missing, this means that the root of the \textit{Tr}, which is the argument $\alpha_1$ is judged as Warranted, and therefore the explanation is untrusted. Otherwise, if all the final four arguments are generated, the \textit{Tr} will be Unwarranted, and the argument $\alpha_1$ will be defeated. 

Let's get back to the point where Altruist chose to hide \textit{W}. If the user was given the whole interpretation of LIME, including \textit{W}, neither argument $\alpha_{6,2,inc}$ nor $\alpha_{6,2,dec}$ would be available to support that the importance of \textit{W} was truthful. This happens because Altruist evaluated this positive importance and found it untruthful for both alterations. Then, if \textit{W} was given to the end user, the claim ($c_2$) could have been raised, and the $\alpha_{5,2}$ argument would have no opponent and would attack $\alpha_{4,2}$, allowing the claim $c_2$ to support $\alpha_{1}$ via $\alpha_{3}$, resulting in an untruthful interpretation. Figure~\ref{untruthTrees} depicts this alternative scenario, considering the elements inside the red boxes as apparent, regarding the third untruthful feature importance score.

\section{Experiments}
\label{s:exp}

In this section, we qualitatively showcase the ability of Altruist to identify untruthful interpretations, followed by an explanation. Moreover, we quantitatively test Altruist on a range of FI techniques. Altruist is evaluated in 3 separate datasets, 3 uninterpretable ML models and 1 interpretable (Table~\ref{tab:datasets}), as well as 4 FI techniques. Nevertheless, the following experiments are not intended to identify the best model or the best FI technique, but to highlight the fact that untruthful features can be identified using Altruist. The experiments' source code will be available in GitHub and DockerHub (\href{Altruist}{\url{https://github.com/iamollas/Altruist}}).

\begin{table}[ht]
    \centering
    \resizebox{0.45\textwidth}{!}{%
    \begin{tabular}{rccc}
                &  Banknote & Heart (Statlog) & Adult Census \smallskip\\
         \hline \noalign{\smallskip}
     instances  & 1372  & 270   & 48842 (1000) \\
     features   & 4     & 13    & 14 (80)    \smallskip\\
         \hline \noalign{\smallskip}
    LR & 98.86\% & 81.21\%  & 94.67\% \\
    SVM & \textbf{100.00\%} & \textbf{81.95\%} & \textbf{95.93\%}  \\
    RF & 99.26\% & 81.89\% & 94.32\% \\
    NN  & 99.43 & 77.02\% & 94.42\% \smallskip\\
    \hline \noalign{\smallskip} \\
    \end{tabular}
    }
    \caption{For each of the 3 datasets: main statistics (top), F$_1$ scores of the 4 different ML models (bottom)}
    \label{tab:datasets}
\end{table}

\begin{table*}[ht]
\centering
\resizebox{\textwidth}{!}{%
\begin{tabular}{c|c|c|c|c|c|c|c|c|c|c|c|c|c|c|c|}
\cline{2-16}
 & LIME & SHAP & PI & IN & AL & LIME & SHAP & PI & IN & AL & LIME & SHAP & PI & IN & AL \\ \hline
\multicolumn{1}{|c|}{LR} & 49.74\% & 37.30\% & \textbf{0.00\%} & \textbf{0.00\%} & \textbf{0.00\%} & 61.71\% & 51.79\% &  \textbf{38.46\%} & \textbf{0.00\%} & \textbf{36.87\%} & 18.54\% & 23.96\% & \textbf{17.18\%} & \textbf{0.00\%} & \textbf{13.56\%} \\ \hline
\multicolumn{1}{|c|}{SVM} & 41.09\% & 56.45\% & \textbf{29.19\%} & - & \textbf{25.91\%} & 59.20\% & 48.60\% & \textbf{46.21\%} & - & \textbf{40.91\%} & 25.23\% & 12.05\% & \textbf{10.79\%} & - & \textbf{10.06\%} \\ \hline
\multicolumn{1}{|c|}{RF} & 91.56\% & \textbf{77.90\%} & 87.54\% & 91.27\% & \textbf{74.00\%} & 76.61\%  & 75.98\% & 84.87\% & \textbf{73.87\%} & \textbf{66.10\%} & 70.01\% & \textbf{13.08\%} & 24.83\% &  62.61\% & \textbf{13.08\%} \\ \hline
\multicolumn{1}{|c|}{NN} & 39.89\% & 47.85\% & \textbf{17.26\%} & - & \textbf{14.43\%} & 58.86\% & \textbf{47.58\%} & 69.23\% & - & \textbf{43.73}\% & 21.51\%  & 26.64\% & \textbf{17.89\%}  & - & \textbf{14.66\%} \\ \hline
 & \multicolumn{5}{c|}{Banknote} & \multicolumn{5}{c|}{Heart (Statlog)} & \multicolumn{5}{c|}{Adult Census} \\ \cline{2-16} 
\end{tabular}%
}
\caption{Percentage of untruthful feature importances per interpretation technique, for the 3 dataset, among the 4 different classifiers. The most truthful technique for each of the models per dataset is denoted with bold. Altruist is the ensemble of the LIME, SHAP and PI techniques. (IN = Intrinsic, AL = Altruist)}
\label{tab:results}
\end{table*}

We are utilising the datasets: Banknote (identify real/fake banknotes) and Heart Statlog (predict absence/presence of a heart's disease)~\cite{ucidata}, and Adult Census~\cite{adultDataset} (predict if income exceeds 50K/yr or not), and the ML models: Logistic Regression (LR), Support Vector Machines (SVM), Random Forests (RF) and Neural Networks (NN). To provide the unbiased performance of each algorithm, 10-fold cross-validation grid searches are performed\footnote{The grid search parameters can be found in GitHub: \url{https://github.com/iamollas/Altruist}. In addition, the optimal set of parameters for each model per dataset, the selection and engineering of features and the undersampling strategies (used in the Adult Census, 1000 instances, 80 features) can also be found in the repo.}. The results are presented in Table~\ref{tab:datasets}. The interpretation techniques selected for this set of experiments are PI, LIME, SHAP, and when available the models' intrinsic interpretation. Only LR and RF can provide intrinsic and pseudo-intrinsic\footnote{A procedure similar to PI, as proposed in the original paper.} interpretations.

\subsection{Qualitative}

For the qualitative experiments, we select the Banknote dataset, due to the small number of features, which makes the example easier to follow. The ML model selected is the SVM, which achieves the perfect F$_1$ score (100\%). The Banknote dataset contains instances with 4 features $F=[f_1,f_2,f_3,f_4]$, where $f_1$ is the \textit{variance}, $f_2$ the \textit{skew}, $f_3$ the \textit{curtosis} and $f_4$ the \textit{entropy}. We did various experiments, two of which are presented here.

\begin{figure}[ht]
    \centering
    \includegraphics[width=0.8\linewidth]{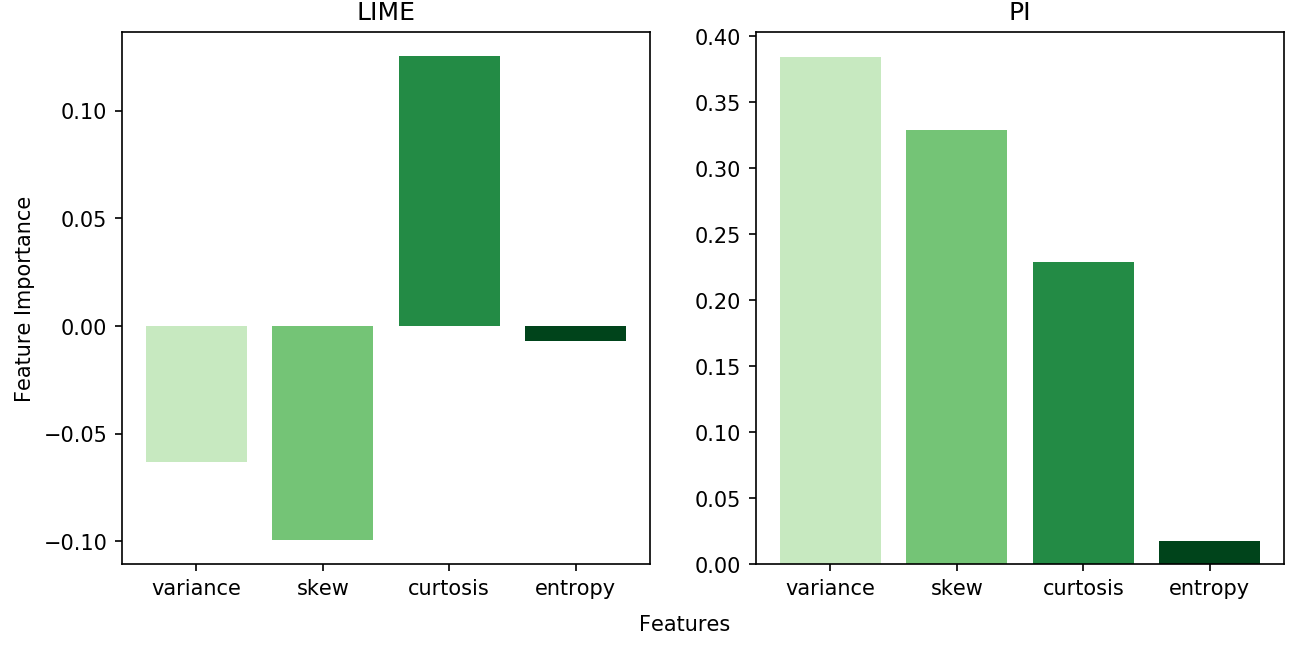}
    \caption{Interpretation of SVM's classification}
    \label{SVM_Inter}
\end{figure}

We select the following random instance: $x_{i}=[0.38, 0.78, 0.76,$ $-0.45]$. The SVM classified the $i_{th}$ banknote as fake with a $81.45\%$ probability. In Figure~\ref{SVM_Inter}, original interpretations provided by LIME and PI for the SVM's prediction are shown. Altruist discovered untruthful importance given to \textit{variance} and \textit{skew} in LIME's result, as well as \textit{entropy's} importance provided by PI. To evaluate the judgement of Altruist, we conducted the following manual experiments. 

LIME assigned a negative importance to the \textit{variance} feature. However, Altruist found it untruthful. We have therefore proceeded to two tests to ensure that Altruist was correct. With respect to the distribution of the \textit{variance} feature, with range $[-7.04, 6.83]$, mean = $0.43$, and std = $2.84$, initially, we increased the value of \textit{variance} from $0.38$ to $0.74$ and the probability increased from $81.45\%$ to $93.49\%$ instead of decreasing. We also altered the \textit{variance's} value from $0.38$ to $0.02$, and the probability decreased from $81.45\%$ to $56.46\%$ instead of increasing. Therefore, Altruist correctly concluded that the importance of this feature was untruthful.

With respect to the distribution of the \textit{entropy} feature, with range $[-8.55, 2.45]$, mean = $-1.19$, and std = $2.1$, initially, we increased the value of \textit{entropy} from $-0.45$ to $-0.08$ and the probability decreased from $81.45\%$ to $71.19\%$ instead of increasing. We also altered the \textit{entropy's} value from $-0.45$ to $-2.23$, and the probability increased from $81.45\%$ to $97.57\%$ instead of decreasing. Hence, Altruist correctly concluded that the importance of \textit{entropy} given from the PI, was untruthful.

\begin{figure}[ht]
    \centering
    \includegraphics[width=0.8\linewidth]{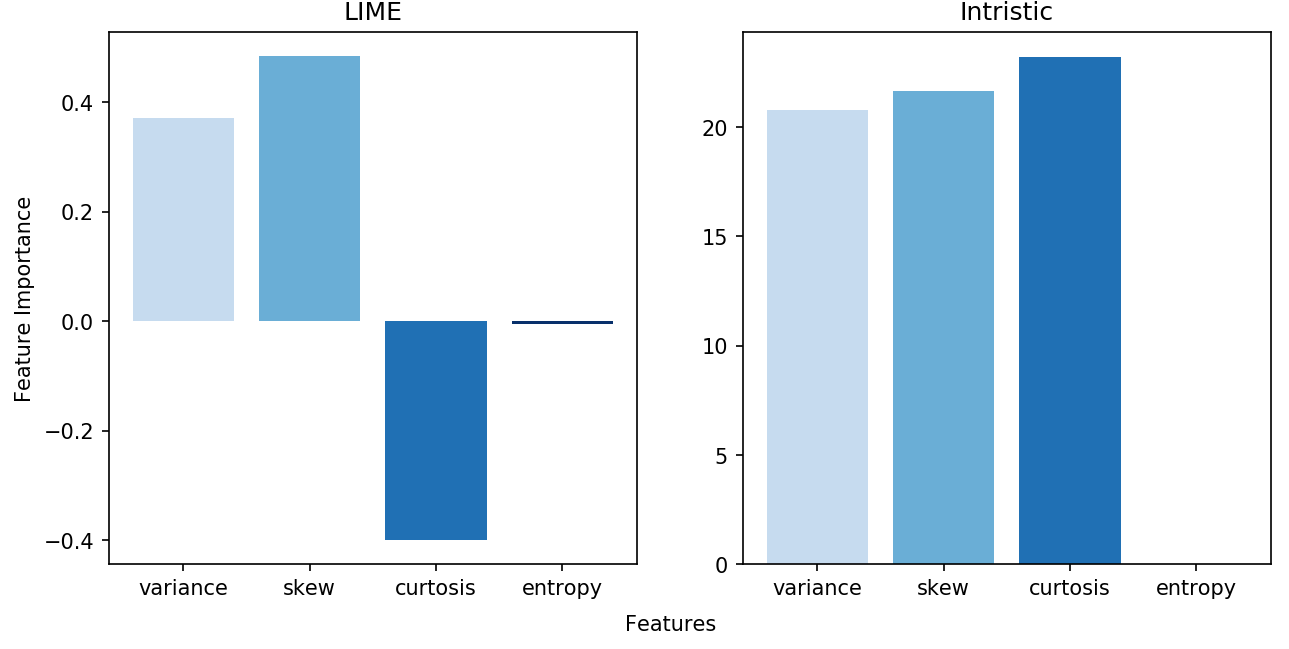}
    \caption{Interpretation of LR's classification}
    \label{LR_Inter}
\end{figure}

We are presenting one more example using the LR model for the same dataset. We select one more random instance: $x_{j}=[-3.38,$ $-13.77, 17.92,  -2.03]$. The LR classified the $j_{th}$ banknote as fake with a $47\%$ probability. In Figure~\ref{LR_Inter}, original interpretations provided by LIME and the actual weights of the LR (intrinsic) are shown. Altruist discovered untruthful importance given to \textit{curtosis} and \textit{entropy} in LIME's result, and no untruthful importances to the intrinsic interpretation, as it was expected considering the interpretable nature of LR. To evaluate the judgement of Altruist, we conducted the following manual experiments. 

We will test the importance given to \textit{curtosis} by LIME. The distribution of the \textit{curtosis} feature has a range of $[-5.29, 17.92]$, mean = $1.4$, and std = $4.31$. We cannot further increase the value of \textit{curtosis} because it already has the maximum value ($17.92$). Thus, we will only test it by decreasing slightly the value from $17.92$ to $16.77$. We see that the probability dropped from $47\%$ to $8.01\%$ rather than increasing, as implied by the positive importance value. Furthermore, we already know that \textit{curtosis} is a significantly influential feature of LR based on its real weights. We may indeed say that Altruist correctly detected LIME's erroneous importance assigned to \textit{curtosis}.

These experiments demonstrate how Altruist attempts to mimic human behaviour to evaluate an explanation. If given such an explanation, a user would, naturally, attempt two alterations to see how the model behaves. It is worth noting that the alterations made by Altruist are indeed small, in terms of the feature's distribution. As seen in the preceding examples, we are slightly altering the feature's value each time.

\subsection{Quantitative}
To quantitatively evaluate Altruist's ability to detect untruthful features, as well as to select the best interpretation technique among many, we will test it in 3 datasets, 3 uninterpretable ML models and 1 interpretable (Table~\ref{tab:datasets}). The results are visible in Table~\ref{tab:results}, describing the mean ratio of untruthful features' importance per interpretation.

\begin{itemize}
    \item[]\textbf{Banknote:} In this case, the SVM achieves the higher F$_1$ ($100\%$). Among the 4 models, LR provides the most truthful interpretations, second comes the NN, and third the SVM. All interpretation techniques struggle to provide truthful explanations for the RF.
    
    \item[]\textbf{Heart (Statlog):} Here, the SVM achieves the higher F$_1$ ($82\%$). Among the 4 models, LR provides the most truthful interpretations, second comes the SVM with PI, and third the NN with SHAP. At the same time, every interpretation technique struggles to provide truthful explanations for the RF.

    \item[]\textbf{Adult Census:} For this dataset, the SVM achieves the higher F$_1$ score ($96\%$). Among the 4 models, LR provides the most truthful interpretations, second comes the SVM with PI technique, and third the RF with SHAP. In contrast to the other 2 test cases, SHAP provides reasonable interpretations for the RF.
\end{itemize}

The effect of Altruist is not mentioned in the aforementioned. Based on the experiments referred to above, we can infer that Altruist is a critical tool for evaluating interpretations given by non-intrinsic techniques such as LIME, SHAP, and PI, detecting on average 43.79\% of untruthful features in 35 out of 36 tests. Altruist can be used effectively as a selection tool (an ensemble), achieving the lowest percent in every case. Provided that no interpretation approach has prevailed over the others, it appears to be an ideal tool for selecting the best technique automatically in a setup where several techniques are used in parallel to interpret an instance's prediction.

In terms of Altruist's scalability, we can state that its performance is proportional to the model's inference time and the response time of the interpretation techniques. However, we measured Altruist's response time, which, on average, evaluates an FI in 0.05 seconds (measured across the 4 models in the 3 datasets).

\section{Conclusion}
IML has emerged as an important research area to interpret ML algorithms. A lot of IML approaches are presenting their interpretations in a feature importance manner. Argumentation is the study of how conclusions can be reached through logical reasoning; that is, claims based, soundly or not, on premises. This paper presents Altruist, a preliminary technique which combines FI interpretation techniques and logic-based argumentation, to provide truthful interpretations on the decision-making of ML models to the end users. It provides the local maximum truthful interpretation, as well as the justification for the truthfulness. Altruist is also presented as an evaluation metric, strongly influenced by the metrics of faithfulness and infidelity. Moreover, it can be used as a tool for automatic selection of the most truthful interpretation among a variety of multiple different interpretations. In future work, the meta-explanation aspect of Altruist is going to be explored. Altruist will also be evaluated in other ML tasks (e.g., regression, multi-label classification) and data types (text, time-series and possibly image). Another aspect that could be investigated is the alteration of categorical feature values, and the effect of the localness of features. Finally, a human-oriented evaluation will be conducted to validate the usefulness of the meta-explanations and the available arguments.

\section*{Acknowledgments}
This paper is supported by the European Union's Horizon 2020 research and innovation programme under grant agreement No 825619 (AI4EU Project).

\bibliographystyle{unsrt}  
%\bibliography{bib}

\end{document}